\pdfoutput=1

\documentclass[11pt]{article}

\usepackage{EACL2023}

\usepackage{times}
\usepackage{latexsym}

\usepackage{graphicx}
\usepackage{amsmath}
\usepackage{algorithm}
\usepackage{algorithmic}

\usepackage[T1]{fontenc}

\usepackage[utf8]{inputenc}

\usepackage{microtype}

\usepackage{multirow}

\usepackage{enumitem}

\usepackage{inconsolata}

\title{The Role of Semantic Parsing in Understanding Procedural Text}

\author{Hossein Rajaby Faghihi$^1$, Parisa Kordjamshidi$^1$, Choh Man Teng$^2$, \and James Allen$^2$ \\
    $^1$ Michigan State University,
  $^2$ Florida Institute for Human and Machine Cognition\\
  \texttt{
    \{rajabyfa, kordjams\}@msu.edu,
    \{jallen, cmteng\}@ihmc.org
    }
    }

\begin{document}
\setlength{\abovedisplayskip}{3pt}
\setlength{\belowdisplayskip}{5pt}
\maketitle
\begin{abstract}
In this paper, we investigate whether symbolic semantic representations, extracted from deep semantic parsers, can help to reason over the states of involved entities in a procedural text. We consider a deep semantic parser~(TRIPS) and semantic role labeling as two sources of semantic parsing knowledge. 
First, we propose PROPOLIS, a symbolic parsing-based procedural reasoning framework.
Second, we integrate semantic parsing information into state-of-the-art neural models to conduct procedural reasoning.
Our experiments indicate that explicitly incorporating such semantic knowledge improves procedural understanding. This paper presents new metrics for evaluating procedural reasoning tasks that clarify the challenges and identify differences among neural, symbolic, and integrated models.   
\end{abstract}

\section{Introduction}
\begin{figure}[h]
    \centering
    \includegraphics[width=\linewidth]{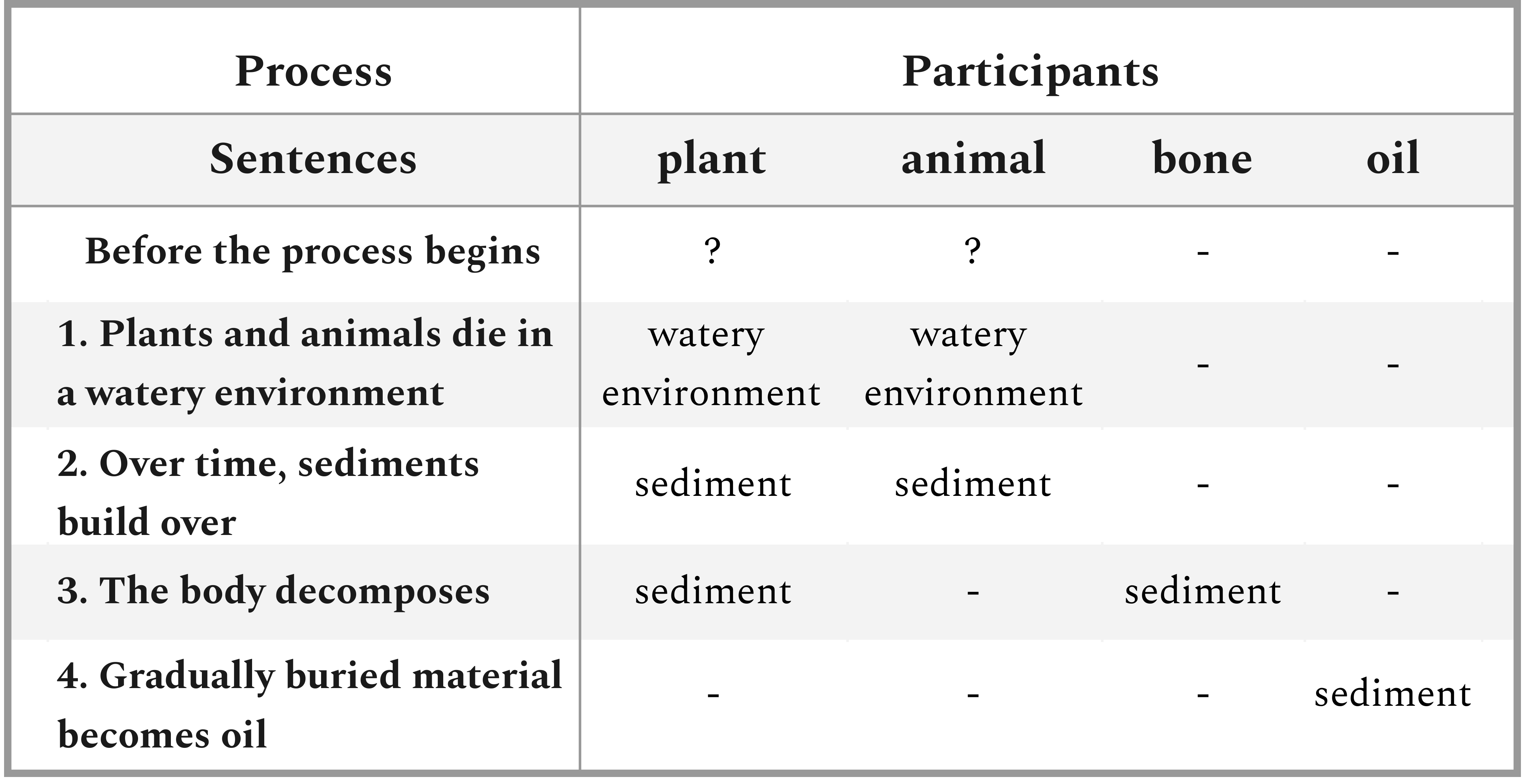}
    \caption{An example of procedural text and its annotation~(location of objects). `-' means the entity does not exist; `?' means the entity's location is unknown. 
    }
    \label{fig:example}
    \vspace{-7mm}
\end{figure}

Procedural reasoning is the ability to track entities and understand their evolution given a sequence of actions~\cite{tandon_etal_2020_dataset}. This kind of reasoning is crucial in understanding recipes~\cite{bosselut2017simulating,yagcioglu2018recipeqa}, manuals and tutorials~\cite{tandon_etal_2020_dataset,wu2022understanding}, cyber-security text~\cite{pal-etal-2021-constructing}, natural events~\cite{tandon_etal_2020_dataset}, and even stories~\cite{storks2021tiered}. An example of a procedural text in the natural event domain, its entities of interest, and their state changes are shown in Figure~\ref{fig:example}.

Inferring actions and their impact on entities involved in a procedural text can be challenging in various aspects. \textbf{First}, there are dependencies between steps to be considered in predicting a plausible action set. For instance, an entity destroyed at step $t$ of the process cannot be moved again at step $t+1$. \textbf{Second}, some sentences contain ambiguous local signals by including multiple action verbs. For example, "The oxygen is consumed in the process of forming carbon dioxide.", where the oxygen is being destroyed, and the carbon dioxide is being created. \textbf{Third}, the sentences are incomplete in some steps. For instance, a step of the process might only indicate "is buried in mud", which cannot be understood without context. \textbf{Fourth}, finding the properties of some entities may require reasoning over both the global context and local relations. For instance, in the sentences ``1. Magma rises to the surface. 2. Magma cools to form lava'', the location of `Lava' after step 2 should be inferred from the prior location of Magma, which is indicated in its previous step. \textbf{Fifth}, common sense is required to understand some consequences. For example, in Figure \ref{fig:example}, step 3, one should use common sense to realize that `decomposing body' would expose the `bones', which will be left behind in the `sediment'. \textbf{Sixth}, understanding some relations requires an advanced co-reference resolution. In Figure~\ref{fig:example}, step 4, a complex co-reference resolution is required to understand that the `buried material' refers to both `plants and animals bones' and that they are transforming into the `oil'.

Except for the common-sense~\cite{zhang2020knowledge} and the ability to make consistent global decisions actions~\cite{gupta2019tracking}, the other challenges might have only been indirectly tackled in the recent research~\cite{huang_etal_2021_reasoning, faghihi2021time}, but have neither been addressed explicitly nor properly evaluated to measure their success on resolving these challenges.
In this paper, we evaluate whether semantic parsers can alleviate some of these challenges.
Semantic parsers provide semantic frames that identify predicates and their arguments in a sentence. For instance, in the sentence `Move bag to the yard', ``Move'' is the predicate, ``bag'' and ``the yard'' are the arguments with types "affected"\footnote{referred to as `Patient' in some other parsing formalisms.} and ``location'' respectively. 
Such semantic information can help disambiguate multi-verb local connections between predicates and arguments~\cite{huang_etal_2021_reasoning}. They can also provide meaningful local relations, making it easier to connect global information to infer entities' states. For instance, in the same sentence, "Magma cools to form lava", "Magma" is noted as the `affected' and `lava' is the result of the predicate `form'. This makes it easier to infer that the location of `lava' should match the last location of `magma'. 

For our study, we consider both the classic semantic role labeling~(SRL)~\footnote{\url{https://demo.allennlp.org/semantic-role-labeling}}, based on \cite{shi2019simple}, which is a relatively shallow semantic parsing model, as well as the deep semantic parser TRIPS\footnote{\url{http://trips.ihmc.us/parser/cgi/parse}}~\cite{ferguson1998trips,allen2017broad}. To investigate the effect of semantic parsing on procedural reasoning, we analyze its effect as a standalone symbolic model as well as its integration in a neuro-symbolic model that combines semantic parsing with state-of-the-art neural models to solve the procedural reasoning task.

First, we design a set of heuristics to extract a symbolic abstraction from the TRIPS parser, called PROPOLIS. We use this baseline to further showcase the effectiveness of semantic parsing information in solving the procedural task.
Next, we integrate the semantic parsers with two well-established procedural reasoning neural backbones, namely NCET~\cite{gupta2019tracking} and TSLM~\cite{faghihi2021time}~(and its extension CGLI~\cite{ma2022coalescing}), through encoding the semantic relations as a graph attention neural network~(GAT)~\cite{shi2020masked}.

For our experiments, we use Propara dataset~\cite{tandon_etal_2020_dataset} that introduces the procedural reasoning task over natural events {that are described in English}.  We realized the {existing evaluation metrics of this dataset} do not reflect the actual performance of the models and fail to identify the challenges and shortcomings of the models. Consequently, we propose new evaluation criteria to shed light on the differences between the models, even when they perform similarly based on the prior metrics.


In summary, our contributions are
(1) Proposing a symbolic model~(Propolis) to solve the procedural reasoning task based on semantic parsing, (2) Proposing a set of new evaluation metrics which can identify the strength and weaknesses of the models, and (3) Showcase the benefits of integrating semantic parsing into the neural models. The code and models proposed in this work are all available in GitHub~\footnote{\url{https://github.com/HLR/ProceduralSemanticParsing}}.

\section{Related Research}
Procedural text understanding has been investigated in many benchmarks such as ScoNe~\cite{long2016simpler}, bAbI~\cite{weston2015towards}, and ProcessBank~\cite{berant2014modeling}. Recent research has focused on procedural reasoning as tracking entities throughout a procedural text. Datasets such as Propara~\cite{tandon_etal_2020_dataset}, Recipes~\cite{bosselut2017simulating}, Procedural Cyber-Security text~\cite{pal-etal-2021-constructing}, and OpenPI~\cite{tandon_etal_2020_dataset} are in the same direction. Procedural reasoning can also be influential in addressing causal reasoning (WIQA)~\cite{tandon2019wiqa}, story understanding~(Trip)~\cite{storks2021tiered}, and abstractive multi-modal question answering~(RecipeQA)~\cite{yagcioglu2018recipeqa}. 

This paper primarily focuses on tracking entities' states and properties throughout a procedural text. Recent research has addressed this problem by predicting actions and properties on local context~(Prolocal)~\cite{mishra2018tracking}, auto-regressive global predictions based on distance vectors~(Proglobal)~\cite{mishra2018tracking}, integrating structural common-sense knowledge built over VerbNet~(ProStruct)~\cite{tandon2018reasoning}, building dynamic knowledge graphs over entities~(KG-MRC)~\cite{das2018building}, explicitly encoding the model to explain dependencies between actions~(XPAD)~\cite{dalvi2019everything}, formulating local predictions and global sequential information flow and sequential constraints~(NCET)~\cite{gupta2019tracking}, formulating the task in a QA setting~(DynaPro, TSLM)~\cite{amini2020procedural,faghihi2021time}, integrating common-sense knowledge from ConcetpNet~(KOALA)~\cite{zhang2020knowledge}, utilizing large generative language models~(LEMON)~\cite{shi2022lemon}, or using both the question answering setting and sequential structural constraints at the same time~(CGLI)~\cite{ma2022coalescing}. All the models mentioned above investigate different neural architectures to tackle the task, while we are more interested in augmenting them with additional knowledge from semantic parsers. 
Recent research has also investigated the integration of semantic role labeling into the procedural reasoning task~(REAL)~\cite{huang_etal_2021_reasoning}, which is very close to our goal in this paper. However, in this work, we propose and investigate a variety of combinations, a deeper semantic representation~(TRIPS) and named relations, in addition to a symbolic approach for solving the procedural reasoning task solely based on semantic parsing.

\section{Technical Approach}
\textbf{Problem definition} The procedural reasoning task can be formally defined by a procedural text including $m$ steps, $S = \{s^1, s^2, ..., s^m\}$, a set of entities $E = \{e_1, e_2, ..., e_n\}$, where $n$ is the number of entities, and a set of properties. Specifically, in the Propara dataset, the property of interest is only the location of the entities $P_L = \{{p_{L}}_{1}^{0}, {p_{L}}_{1}^{1}, ..., {p_{L}}_{n}^{m}\}$, where ${p_L}_i^t$ denotes the $j$th entity at step $t$. In Propara, the location prediction starts at step $0$, which indicates the entity's location before the process begins. The location of an entity can either be known~(represented by a string) or unknown~(represented by "?"). 
Similar to prior research~\cite{tandon_etal_2020_dataset}, the location property is used to infer a set of actions 
$A = \{a_1^1, a_2^1, .., a_n^m\}$, where $a_t^j$ denotes the action type applied to entity $j$ at step $t$.
 Following the prior research~\cite{mishra2018tracking}, we extract all the noun phrases from the sentences and only consider those as location candidates.

\begin{figure}[h]
    \centering
    \includegraphics[width=0.6\linewidth]{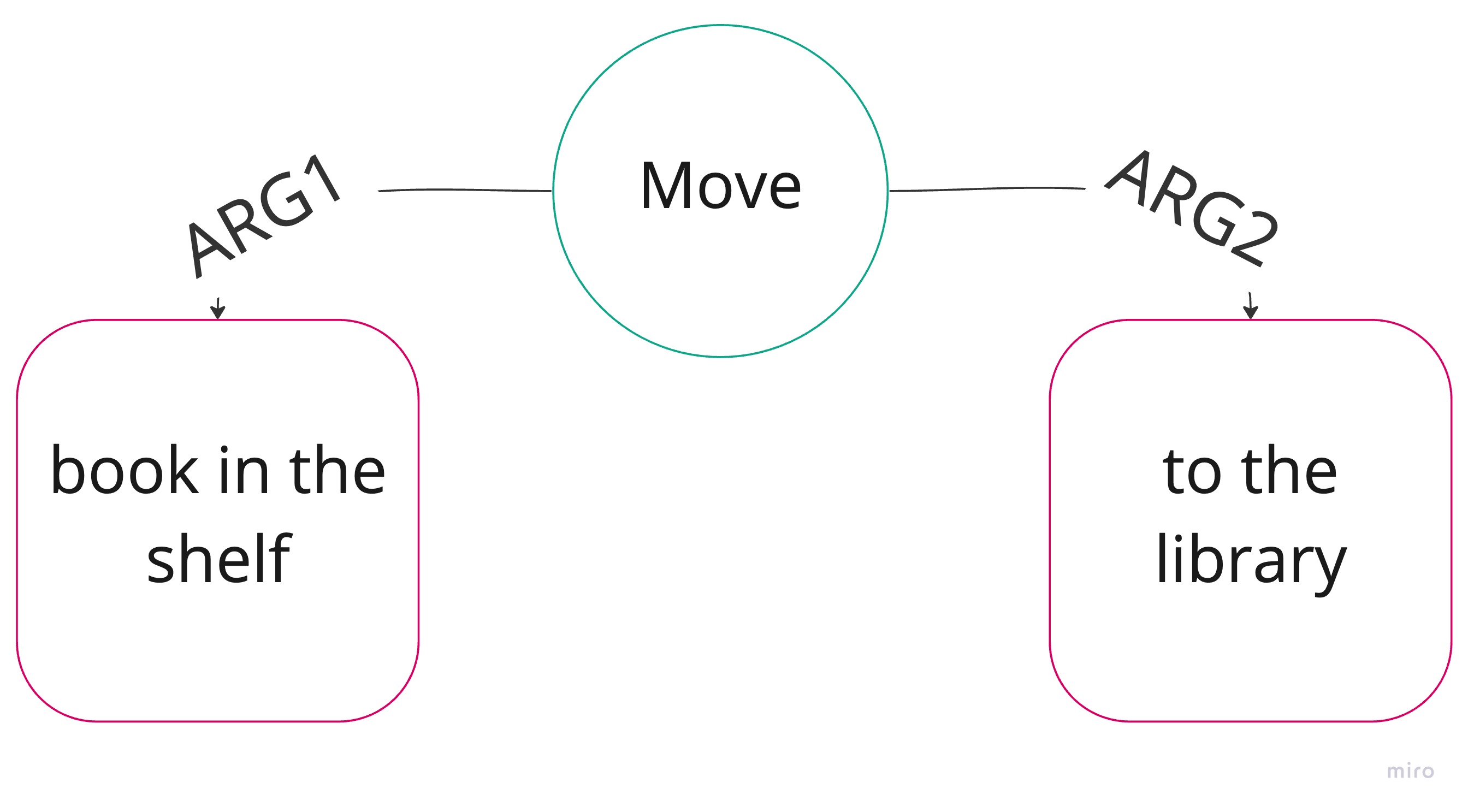}
    \caption{The SRL annotation for the sentence ``Move the book in the shelf to the library''.}
    \label{fig:srl}
    \vspace{-4mm}
\end{figure}

\begin{figure}[h]
    \centering
    \includegraphics[width=0.8\linewidth]{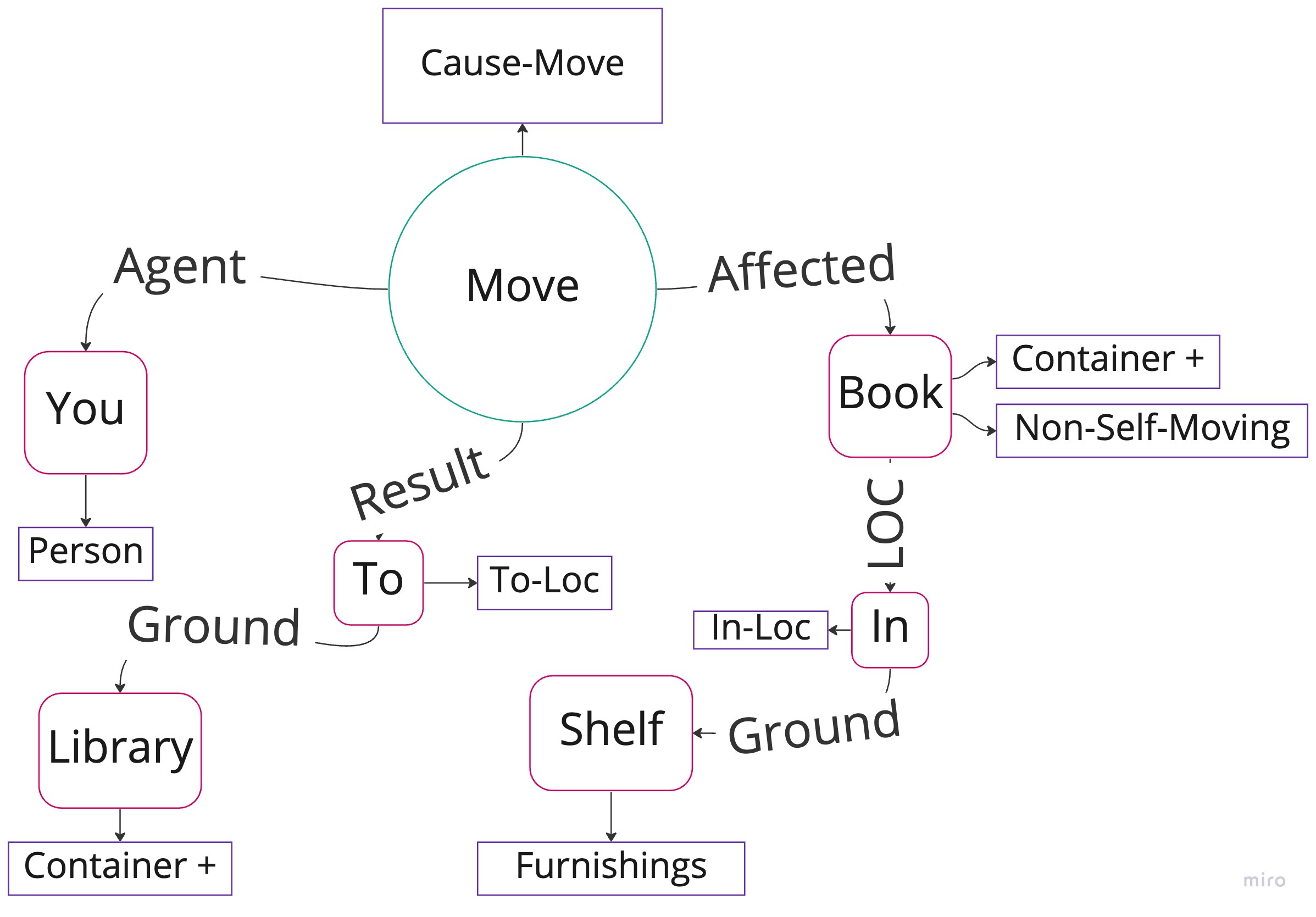}
    \caption{The TRIPS parse for the sentence ``Move the book in the shelf to the library''.}
    \label{fig:trips}
    \vspace{-3mm}
\end{figure}

We investigate two different modeling approaches to solve this problem. First, we use a symbolic and parsing-based model, and second, we integrate semantic parsing with neural models.
We use two different sources for semantic extraction: SRL and TRIPS.
In general, SRL is coarse-grained and shallow compared to TRIPS. The connections in TRIPS are not limited to the pairwise connections between predicates and arguments but are extended to the semantic connections between any two words.
Since TRIPS relies on a general purpose ontology, it also augments the arguments and predicates with additional information about a set of possible features~(mobility, container, negation) and mapping of the words to hierarchical ontology classes~(i.e., mapping ``water'' to ``beverage'').
SRL is centered around the semantic frames of the verbs (~predicates) and identifies each predicate's main and adjunct~(mainly time and location) arguments in the sentence.
Figure \ref{fig:srl} and \ref{fig:trips} show examples of the SRL and TRIPS parses, respectively.


The symbolic model only uses the TRIPS parser as it provides more extended extractions and meaningful relations, while both SRL and TRIPS are used for integration with the neural baselines.


\subsection{PROPOLIS: Symbolic Procedural Reasoning}
We propose the PROPOLIS model, which solves the procedural reasoning task merely by symbolic semantic parsing. PROPOLIS operates on the TRIPS parser in three steps. First, it makes an abstraction over the original parse to summarize the information in the graph and include a smaller set of actions and changes in objects and their locations. Second, it uses a set of rules to transform the abstracted parses into clear actions and identifies the affected objects by the actions, using the semantic roles, while extracting an ending location or starting location. Lastly, it performs global reasoning to connect the local decisions and produce a consistent sequential set of actions/locations for each entity of interest. More details about the steps are also available in Appendix \ref{appx:propolis}.

\subsubsection{Graph Abstraction} 

The original TRIPS parse includes many concepts and edges that do not directly affect entities' location or existence. Therefore, we make a more concise graph abstraction to facilitate processing the entities, actions, and locations. 
To obtain a more informative abstraction, firstly, the relevant classes of the TRIPS ontology are mapped to action classes defined in the Propara dataset~(Create, move, or destroy). For instance, the verb `flow' is first mapped to the `fluidic motion' class in the TRIPS ontology, which is a child of the `motion` class, and the `motion' class is mapped to the `move' action in the Propara dataset. This will help distinguish the predicates that signal a change in the location or existence of objects. 
Second, the important arguments are identified in the parse, and the locations are extracted. 
The graph is decomposed to include a set of events with their arguments. Each event may contain different roles such as "agent", "affected", "result", "to\_location", "from\_location", or other roles required by its semantic frame.
\begin{table*}[t]
    \centering
    \begin{tabular}{|c|c|c|}
    \hline
         Main Predicate & Roles & Decisions \\ \hline
         Move& Affected, Agent &  The ``Affected'' is being moved. \\ \hline
         Move & Agent & The ``Agent'' is being moved. \\ \hline
         Destroy & Affected & The ``Affected'' is being destroyed \\ \hline
         Create & Affected\_Result, Affected & The ``Affected\_Result'' is being created \\ \hline
         Create & Affected & The ``Affected'' is being created \\ \hline
         Change & Affected, Res & \shortstack{The ``Affected'' is being destroyed, and \\the ``Res'' is being created} \\ \hline
    \end{tabular}
    \caption{The rules used to evaluate the effect of actions on various roles of the semantic frame}
    \label{tab:heuristics}
    \vspace{-4mm}
\end{table*}
\subsubsection{Rule-based Local Decisions}
We use a set of heuristic rules to map the abstracted graph onto actual actions over the entities of interest. The rules are written according to the semantic frames and the type of  predicates and arguments in each parse. For instance, if a semantic frame is mapped to `Move' and has both the `agent' and `affected' arguments, then the `affected' argument specifies the object being moved. The same frame with only an `agent' argument indicates a move for the object in the `agent' role. {Table \ref{tab:heuristics} shows the most frequent templates we used to transform the local parses into actual decisions over the entities.

\subsubsection{Global Reasoning}
The two first steps are merely based on the local sentence-level actions of each step. 
We need additional global reasoning over the whole procedure to predict the outputs. 
Global reasoning ensures that local decisions form a valid global sequence of actions for a given entity. For instance, if an entity is predicted to be destroyed at step 2 and moved at step 3, we consider the `destroyed' action a wrong local decision since a destroyed object cannot move later in the process. The graph also contains passive indications of object location in phrases such as "the book on the shelf" or even indications of prior locations in terms of a `from\_location' argument. These phrases do not generate actions but provide information that should be used in previous steps. For example, if step $t$ has a local prediction `Move' for entity $e$ with no target location and step $t+1$ has a `from\_location' for entity $e$, then the `from\_location' should be used as the target location of the `Move' action in the previous step. 




\subsection{Integration with Neural Models}
\label{sec:integration}
Here, we investigate whether explicitly incorporating semantic parsers with neural models can help better understand the procedural text. We choose two of the recently proposed and most commonly used backbone architectures for procedural reasoning tasks, namely NCET~\cite{gupta2019tracking} and TSLM~\cite{faghihi2021time}~(and its extension CGLI~\cite{ma2022coalescing}). Similar to \cite{huang_etal_2021_reasoning}, we rely on a graph attention network~(GAT) to integrate the information from the semantic parsers into the neural baselines.

Following~\cite{huang_etal_2021_reasoning}, the nodes in this graph are either (1) predicates in the semantic frames, (2) mentions of entities of interest~(Exact match or Co-reference), or (3) noun phrases in the sentence. An edge in the SRL graph exists between two nodes if they have a (predicate, argument) connection or they are both parts of the same verb semantic frame~(argument to argument)~\cite{huang_etal_2021_reasoning}.
It is relatively straightforward to build a semantic graph with the TRIPS parser because it outputs the parse as a graph.

An edge is created between any pairs of nodes~(phrases) in the graph if any subsets of these two phrases are connected in the original parse. The edge types are preserved. Since not all the nodes in the original parse are present in the new simplified graph, we may lose some key connections. 
To fix this, if two nodes~(phrases) are not connected in the new graph but have been connected in the original one, we find the shortest path between them in the original parse and connect them with a new edge with the type being the concatenation of all the edge types in the path. 
Lastly, nodes are connected across sentences based on either an exact match or co-reference resolution. 

Both NCET and TSLM models are trained based on Cross Entropy to compute the loss for both actions and locations. The final loss of the model is calculated by $L_{total} = L_{action} + \lambda * L_{location}$,
where $\lambda$ is a balancing hyper-parameter.



\subsubsection{Integration with NCET as Backbone}
\label{sec:int_ncet}
The NCET model uses a language model to encode the context of the procedure and compute representations for mentions of entities, verbs, and locations. These representations are used in two sub-modules for predicting the actions and the locations. To integrate the semantic parsers with the NCET architecture, we use the output of the language model to initialize the semantic graph representations. Then multiple layers of graph attention network~(based on TransformerConv~\cite{shi2020masked}) are applied to encodes the graph structure. We combine the updated graph representations with the initial mention representations. These combined representations are later used in subsequent prediction modules.

More formally, we start by using a language model to encode the context of the process $h^\prime = LM(S)$, where $S$ is the procedure, and $h^\prime$ is the embedding output from the language model. The representations are further encoded by a BiLSTM $h = BiLSTM(h^\prime)$. 

\noindent\textbf{Graph Attention Network}
Since each node in the semantic graph corresponds to a subset of tokens in the original paragraph, we use the mean average of these tokens' representation to initialize the nodes' embedding denoted as $v^0_i$. If the graph contains edge types, the edges between each two nodes $i$ and $j$ are denoted by $e_{ij}$ and is represented by the average token embedding through the same LM model used for encoding the story, $e_{ij} = Mean(LM(e_{ij}^{text}))$. Lastly, we use $C$ layers of TransoformerConv~\cite{shi2020masked} to encode the graph structure. More details on the graph encoder is available in Appendix \ref{appx:gat}.

\noindent\textbf{Representing Mentions}
To integrate the semantic parses with the baseline model, we use both representations obtained from the language model and the graph encoder to represent entities, verbs, and locations in the process. Mention representations are denoted by 
$r^m_t = \left[M(h^m_t);M({h_g}^m_t))\right],$
where $t$ is one step of the process, $h^m_t$ is the average representation of tokens in the story corresponding to the mention $m$ in step $t$, ${h_g}^m_t$ is the average embedding of nodes corresponding to mention $m$ in step $t$, and the function $M$ replaces the representations with zero if there is no mention of $m$ in step $t$. 

\noindent\textbf{Location Prediction}
We first encode the pairwise representation of an entity $e$ and location candidate $lc$ at each step $t$, denoted by $
\mathbf{x}_t^{(e, lc)}= \left[\mathbf{r}_t^e ; \mathbf{r}_{t}^{lc}\right]$.
Next, we use an LSTM to encode the step-wise flow of the pair representation to get $\bar{h}^{(e, lc)}_t = LSTM(\left[\mathbf{x}_t^{(e, lc)}\right])$. Finally, the probability of each location candidate $lc$ to be the location of entity $e$ at step $t$ is calculated by a $softmax$ over the potential candidates, $p^{(e, lc)_t} = Softmax(W^t_{loc} \bar{h}^{(e, lc)}_t)$, where $W^t_{loc}$ is the learning parameters of a single multi-layer perceptron.

\noindent\textbf{Action Prediction}
To predict the action for entity $e$ at step $t$, we create a new representation for the entity based on its mention and the sentence verbs, denoted by
$x^e_t = \left[r^e_t;Mean_{v \in np(e_t)}(r^v_t)\right],$
where $np(e_t)$ is the set of verbs whose corresponding node in the graph has a path to any of the nodes representing entity $e$ in step $t$. The final representations are then produced using a BiLSTM over the steps, $h^e_t = LSTM([x^e_t])$. Lastly, a neural CRF layer is used to consider the sequential structure of the actions by learning transition scores during the training of the model~\cite{gupta2019tracking}. The set of possible actions is shown in Table \ref{tab:tags}.
\begin{table}[]
\small
    \centering
    \begin{tabular}{|c|c|}
    \hline
         Tag& Description  \\ \hline
         O\_D& Entity does not exist after getting destroyed \\ 
         O\_C & Entity does not exist before getting created \\
         E & Entity exists and does not change \\
         C & Entity is created \\
         D & Entity is destroyed \\ \hline
    \end{tabular}
    \caption{The list of output tags/actions}
    \label{tab:tags}
    \vspace{-5.8mm}
\end{table}

\begin{table*}[]
\centering
\begin{tabular}{|c|c|c|cc|c|c|}
\hline
\multirow{2}{*}{} & Local & Global Loc & \multicolumn{2}{c|}{Global Ent} & Global Loc and Ent & Ambiguous \\ \cline{2-7} 
                  & Both  & Both       & Actions       & Locations       & Both               & Actions   \\ \hline
Train             & 885   & 367        & 438           & 340             & 114                & 593       \\
Dev               & 116   & 44         & 66            & 3               & 9                  & 76        \\
Tests             & 105   & 61         & 98            & 71              & 18                 & 110       \\ \hline
\end{tabular}
\caption{The number of decisions per category of evaluation with the new decision-level metric. ``Both'' refers to both location and action decisions and is used since the number of those decisions is the same in most cases. The number of decisions in the `Global Ent' case can be different for the actions and the locations because this category also considers `destroy' events that  have no corresponding locations.}
\label{tab:eval_stat}
\vspace{-5mm}
\end{table*}

\subsubsection{Integration with TSLM as Backbone}
The TSLM~\cite{faghihi2021time} model reformulates the procedural reasoning task as a question-answering problem. The model simply asks the question, `Where is entity $e$?' at each step of the process. To include the context of the whole process when asking the same question at different steps, TSLM further introduces a time-aware language model that can encode additional information about the time of events. Given the new encoding, each step of the process is mapped to either past, present, or future. TSLM uses the answer to the question at each step to form a sequence of decisions over the location of entity $e$. To integrate the semantic graph with this model, we first extend the graph by adding a question node. The graph is then initialized using the time-aware language model. The encoded representations of the graph, after applying multiple layers of GAT, are combined with the original token representations and used for extracting the answer to the question. 

\noindent\textbf{Initial Representation}
For each entity $e$ and timestamp $t$, the string ``where is e? s1 </s> s2 </s> ... $s_m$ </s>'' is fed into the time-aware language model.  
Accordingly, the tokens' representations for timestamp $t$ are $(h_e)^i_t = LM(S, t)$.


\noindent\textbf{Graph Attention Module}
Inspired by ~\cite{zheng2020srlgrn}, we add new nodes to the semantic graph to represent the question and each step of the process. We connect the question node to any node in the graph representing the entity of interest $e$, and each step node to all the tokens in their corresponding sentence. An example of the QA-based graph can be found in Appendix \ref{appx:sparser}.
All the node embeddings are initialized by the average embedding of their corresponding tokens in the procedure.  
We use $C$ layers of graph attention network~(TransformerConv), similar to Section \ref{sec:int_ncet}, to encode the graph structure. 

\noindent\textbf{Location Prediction}
For predicting the locations of entities, that is, the answer to the question, we predict the answer among the set of location candidates. This is different from the common practice of predicting start/end tokens. We represent each location candidate by combining representations from both the graph and the time-aware language model, denoted by 
$r^{lc}_t = \left[(h_e)^{lc}_t,(h_g^cl)^{lc}_t\right]$, where $(h_g^cl)^{lc}_t$ is the representation of the $lc$ from the last layer of the GAT.
The answer is then selected by calculating a $softmax$ over the set of location candidates, $p^{lc}_t = Softmax(W^{location} r^{lc}_t)$.

\noindent\textbf{Action Prediction}
Similar to CGLI~\cite{ma2022coalescing} model, we  explicitly predict the actions of entities alongside the locations.
First, the model extracts each timestamp's ``CLS'' tokens and builds sequential pairs of $(CLS^e_t, CLS^e_{t+1})$. Then, it produces a change representation vector for each of these pairs, denoted by $r^{e}_t = F(CLS^e_t; CLS^e_{t+1})$. Lastly, the sequence of $[r^e_t]$ logits is passed through the same neural CRF layer used by the NCET model, introduced in Section \ref{sec:int_ncet}, to generate the final probability of actions. 

\section{Evaluation}
\label{sec:evaluation}
We use three evaluation metrics to analyze the performance of the symbolic, sub-symbolic, and neural baselines. The first metric is sentence-level and proposed in~\cite{mishra2018tracking}. The second metric is a document-level evaluation proposed by~\cite{tandon2018reasoning}. Both of these metrics evaluate higher-level procedural concepts that can be inferred from the predictions of the model rather than the raw decisions. These  metrics give more importance to the actions compared to the location decisions. Although they can successfully evaluate some aspects of the models, they fail to measure the research progress in addressing the challenges of the procedural reasoning task. We extend these evaluations with a new decision-level evaluation metric that considers almost all model decisions with a similar weight and evaluates the models based on the difficulty of the reasoning process. 

\subsection{Propara Evaluation Metrics}
With the \textbf{sentence-level} metrics, the predictions are evaluated in three different categories. \textbf{(Cat1)} evaluates whether an entity $e$ has been created~(destroyed/moved) during the process. \textbf{(Cat2)} evaluates when an entity $e$ is created~(destroyed/moved). \textbf{(Cat3)} evaluates where $e$ is created~(destroyed/moved). 

With the \textbf{document-level} metrics, we evaluate the Inputs, Outputs, Conversion, and Moves separately and average over the F1 score of these four criteria to output one F1 score as the final metric. Here, \textbf{Inputs} are entities that did exist before the process started and are destroyed during. \textbf{Outputs} are entities that did not exist before the process but created during it. \textbf{Conversions} evaluates which entities converted to another entity. Lastly, \textbf{Moves} evaluates which entities have been moved from one place to another.

\subsection{Extended Evaluation}
Both sets of existing evaluation metrics of the Propara dataset do not directly evaluate the predictions of the model but rather evaluate higher-level procedural concepts which can be inferred from the sets of decisions~(i.e, an entity being input/output).
Given their evaluation criteria, one model may surpass another in the number of correct decisions but still obtain a lower performance.

Therefore, we propose a new evaluation metric~(\textbf{decision-level}) that directly evaluates the models' decisions. This evaluation metric is designed to consider the difficulty of the reasoning process and help better identify the core challenges of the task.  
We divide the set of decisions into five categories based on the presence of the entity $e$ and the location $l$ at each step $t$. We denote any mention of $e$ by $m^e$, any mention of $l$ by $m_l$, the action for entity $e$ at step $t$ by $tag^e_t$, and the text of the current step by $S_t$.  
The following specifies the five categories and how a decision falls under them.\\
\noindent\textbf{Local Decision}: A decision where (1) $m^e \in S_t$, (2) $m^l \in S_t$, and (3) $tag^e_t \in \{Move, Create\}$ \\
\textbf{Global Location Decision}: A decision where (1) $m^e \in S_t$, (2) $m^l \not\in S_t$, and (3) $tag^e_t \in \{Move, Create\}$ \\
\textbf{Global Entity Decision}: A decision where (1) $m^e \not\in S_t$, (2) $m^l \in S_t$ or $l = "-"$, and (3) $tag^e_t \in \{Move, Create, Destroy\}$ \\ 
\textbf{Global Entity and Location Decision}: A decision where (1) $m^e \not\in S_t$, (2) $m^l \not\in S_t$, and (3) $tag^e_t \in \{Move, Create\}$ \\
\textbf{Ambiguous Local Action}: A decision where (1) $m^e \in S_t$ and (2) $S_t$ contains multiple action verbs. 

{Table \ref{tab:eval_stat} shows the detailed statistics of the number of decisions falling under each of these five categories for the Propara dataset}. Evaluating the performance of models given the new decision-level metric will clarify the lower-level challenges in the reasoning over states and locations of entities simultaneously. Getting accurate predictions in any of these categories of decisions requires the models to have different reasoning capabilities. 

The local decisions mostly require a sentence-level understanding of the action and its consequences. The global location decisions require reasoning over the current step and the ability to connect the local information to the global context. The predictions for the category of the global entity mostly require reasoning over complex co-references~(we have already considered simple co-references such as pronouns as mentions of the entity) or the ability to recover missing pronouns in a sentence such as "Gradually mud piles over (them)". The global entity and location decisions are the most challenging cases, which require reasoning over local and global contexts, complex co-reference resolution, and handling of missing pronouns. The ambiguous decisions mainly require local disambiguation of (entity, role, predicate) connections when multiple predicates are present in the sentence. Moreover, common sense is required for a subset of all the decision categories.  
\begin{table*}[t]
\centering
\small
\begin{tabular}{|c|c|ccccc|ccc|}
\hline
\multirow{2}{*}{\#Row} & \multirow{2}{*}{Models}  & \multicolumn{5}{c|}{Sentence-level evaluation}                                    & \multicolumn{3}{c|}{Document-level evaluation} \\ \cline{3-10} 
                       &                          & Cat1           & Cat2           & Cat3           & Macro-avg     & Micro-avg      & Precision      & Recall       & F1             \\ \hline
1                      & ProLocal                 & 62.7           & 30.5           & 10.4           & 34.5          & 34.0           & \textbf{77.4}  & 22.9         & 35.3           \\
2                      & ProGlobal                & 63             & 36.4           & 35.9           & 45.1          & 45.4           & 46.7           & 52.9         & 49.4           \\
3                      & KG-MRC                   & 62.9           & 40             & 38.2           & 47            & 46.6           & 64.5           & 50.7         & 56.8           \\ \cline{2-10} 
4                      & PROPOLIS(ours)           & 69.9           & 37.71          & 5.6            & 37.74         & 36.67          & 70.9           & 50.0         & 58.7           \\ \cline{2-10} 
5                      & NCET (re-implemented)    & 75.54          & 45.46          & 41.6           & 54.2          & 54.38          & 68.4           & 63.6         & 66             \\
6                      & REAL(re-implemented)$^*$     & 78.9           & 48.31          & 41.62          & 56.29         & 56.35          & 67.3           & 64.9         & 66.1           \\ \cline{2-10} 
7                      & NCET + SRL(ours)         & 77.1           & 46.35          & 42             & 55.16         & 55.32          & 67.8           & 65.2         & 66.5           \\
8                      & NCET + TRIPS(ours)       & 77.1           & 48.12          & 43.36          & 56.19         & 56.32          & 72.5           & 65.4         & 68.8           \\
9                      & NCET + TRIPS(Edge)(ours) & 75.68          & 47.6           & 45.71          & 56.33         & 56.37          & 69.9           & 65.5         & 67.6           \\
10                     & NCET + PROPOLIS(ours)$^+$   & 78.54          & 48.69          & 44.26          & 57.16         & 57.31          & 74.6           & 65.8         & 69.9           \\ \cline{2-10} 
11                     & DynaPro                  & 72.4           & 49.3           & 44.5           & 55.4          & 55.5           & 75.2           & 58           & 65.5           \\
12                     & KOALA                    & 78.5           & 53.3           & 41.3           & 57.7          & 57.5           & 77.7           & 64.4         & 70.4           \\
13                     & TSLM                     & 78.81          & 56.8           & 40.9           & 58.83         & 58.37          & 68.4           & 68.9         & 68.6           \\ \cline{2-10} 
14                     & CGLI                     & 80.3           & \textbf{60.5}  & 48.3           & \textbf{63.0} & \textbf{62.7}  & 74.9           & \textbf{70}  & \textbf{72.4}  \\
15                     & CGLI + TRIPS (ours)      & \textbf{80.62} & 58.94          & \textbf{49.08} & 62.88         & 62.68 & 74.5           & 68.5         & 71.4           \\ \hline
\end{tabular}
\caption{The table of results based on sentence-level and document-level evaluation of the Propara Dataset. $^*$ Since the code for the REAL model is not available, we have re-implemented the architecture based on the guidelines of the paper and the communications. $^+$ The graph is first abstracted using the PROPOLIS graph abstraction phase and then used instead of the Trips parse as input to the model.}
\label{tab:main_res}
\vspace{-1.5mm}
\end{table*}
\begin{table*}[]
\centering
\small
\begin{tabular}{|c|ccc|ccc|ccc|ccc|c|}
\hline
\multirow{2}{*}{Model} & \multicolumn{3}{c|}{Local}                    & \multicolumn{3}{c|}{Global Loc}               & \multicolumn{3}{c|}{Global Ent}               & \multicolumn{3}{c|}{Global Loc and Ent}       & Amb$^+$    \\ \cline{2-14} 
                       & A             & L             & Both          & A             & L             & Both          & A             & L             & Both          & A             & L             & Both          & A             \\ \hline
KOALA                  & 74.3          & 65.7          & 59.0          & \textbf{86.9} & 24.6          & 22.9          & 1.0           & 7.0           & 0.0           & 5.6           & 11.1          & 0             & 73.63         \\
PROPOLIS               & 55.2          & 19.0          & 19.0          & 63.9          & 1.6           & 1.6           & 0.0           & 9.9           & 0.0           & 0.0           & 0.0           & 0.0           & 52.7          \\
NCET                   & 69.5          & 62.8          & 60.0          & 70.5          & 36.1          & 29.5          & 3.1           & 5.6           & 0.0           & 0.0           & 0.0           & 0.0           & 57.2          \\ \hline
NCET + SRL             & 68.6          & 65.7          & 61.9          & 77.0          & 36.1          & 31.1          & 10.2          & 5.6           & 0.0           & 5.5           & 5.5           & 0.0           & 62.7          \\
NCET + TRIPS           & 71.4          & 67.6          & \textbf{63.8} & 75.4          & 42.6          & 36.1          & 10.2          & 9.9           & 2.8           & 5.5           & 11.1          & 0.0           & 63.6          \\
NCET + PROPOLIS        & 71.4          & 64.8          & 61.9          & 83.6          & 36.1          & 34.4          & 3.1           & 7.0           & 0.0           & 5.5           & 5.5           & 0.0           & 70.9          \\ \hline
CGLI                   & 65.7          & 62.9          & 54.3          & 75.4          & 59.0          & 50.8          & \textbf{19.4} & 19.7          & 11.3          & 22.2          & \textbf{27.8} & 11.1          & 70.0          \\
CGLI + TRIPS           & \textbf{75.2} & \textbf{70.5} & 61.9          & 80.3          & \textbf{60.6} & \textbf{52.2} & 17.3          & \textbf{22.5} & \textbf{12.7} & \textbf{27.8} & \textbf{27.8} & \textbf{16.7} & \textbf{74.5} \\ \hline
\end{tabular}
\caption{The results of the models on the new extended evaluation metric~(decision-level) in terms of accuracy~(\%). `A' means the action is correct, `L' means the location is correct, and `Both' means both the action and location are correct.$^+$ Local ambiguous cases.}
\label{tab:extended}
\vspace{-5mm}
\end{table*}
\section{Experiments}
Here, we summarize the performance of strong baselines compared with the symbolic~(PROPOLIS) and integrated models. The implementation details of the models are available in Appendix \ref{appx:imp}. 
Table \ref{tab:main_res} shows the performance of models in the two conventional metrics of the Propara dataset, and Table \ref{tab:extended} shows the performance of models based on the decision-level metric. We summarize our findings in a set of question-answer pairs.

\noindent\textbf{Q1. Can semantic parsing alone solve the problem reasonably?} Based on Table \ref{tab:main_res}, the PROPOLIS model outperforms many of the neural baselines~(document-level F1-score of row\#4 compared to rows \#1 to \#3), showing that deep semantic parsing can provide a general solution for the procedural reasoning task to some extent without the need for training data. This model performs relatively well on action-based decisions~(cat1) but fails to extract the proper location decisions~(cat3). This is because many locations are inferred based on common sense rather than the verb semantic frames. Notably, the set of rules written on top of PROPOLIS is local and simple and can be further expanded to improve performance. Table \ref{tab:extended} further indicates that the predictions of the PROPOLIS model on the actions are much closer to the SOTA models than its predictions for the entities' location. The good performance of PROPOLIS on the action decisions for the ``Global Location'' category can further show that the local context can mostly indicate the action even if retrieving the result of the action~(location) requires more reasoning steps. Lastly, since PROPOLIS is a model built over local semantic frames, it dramatically fails to make accurate decisions when the entity does not appear in the sentence~(Global Ent).

\noindent\textbf{Q2. Can the integration of semantic parsing improve the neural models?}  We evaluate this based on the two strong baselines, NCET and TSLM.
When semantic parsers are integrated into NCET, all three evaluation metrics improve (compare rows \#7 to \#10 with row \#5). This improvement is even better if the source of the graph is the abstracted parse from the PROPOLIS method~(row \#10). Semantic parsers improve NCET's performance in all categories of decisions, particularly in local ambiguous sentences and decisions requiring reasoning over global locations. Notably, the integration of PROPOLIS with the NCET model significantly boosts the ability to disambiguate local information in sentences with multiple action verbs.

The integration of the semantic graph slightly hurts the performance of the CGLI baseline when using conventional metrics~($1\%$). However, it outperforms this baseline on ``cat3''~($0.78\%$), which is the only evaluation that directly considers location predictions.
Notably, the original CGLI model~(baseline) uses the pre-trained classifiers from SQUAD~\cite{rajpurkar2016squad} to predict the start/end tokens from the paragraph as the locations~(answer to the question). However, since the integrated method extracts candidates from the graph in the form of spans, it cannot reuse the same pre-trained classifier parameters. This may contribute to the drop in performance since CGLI performs 2\% lower on the document-level F1 score when SQUAD pre-training is removed~\cite{ma2022coalescing}. Despite the drop in performance based on the conventional metrics, the integrated QA-model~(CGLI + TRIPS) outperforms the baseline in almost all the criteria in the new evaluation~(Table \ref{tab:extended}), especially on decisions that only require local reasoning or local disambiguation. This is due to the global nature of the TSLM~(or CGLI) backbone, which predicts the locations based on the whole story and ignores many of the local signals, whereas the graph can help directly extract the local relations.

\noindent\noindent\textbf{Q3. How can the decision-level metrics help understand models' weaknesses and strengths?} Based on the results in Table \ref{tab:extended}, the NCET model is better at reasoning over the local context than the global context. It also clarifies that although the TSLM~(or CGLI) model can properly reason over multiple steps, it is not as competitive as the NCET model in the local cases. However, the integration of semantic parsers could improve the models to close the gap on both local and global aspects and has a complimentary influence on the initial performance of the baselines. As a general conclusion based on our new evaluation metrics, we can argue that the most challenging decisions are the ones that require reasoning over missing mentions of entities in the local context. Addressing this challenge may require external reasoning over common-sense, performing the complex co-reference resolution, or handling missing pronouns. 

\section{Discussion}
Here, we discuss some of the potential concerns that may arise with the usage of symbolic systems such as TRIPS and the new evaluation criteria.

\noindent\textbf{Coverage and rule crafting of PROPOLIS.} Our implementation of the symbolic method and the integrated models rely on the knowledge extracted from very fine-grained semantics covered in TRIPS. Consequently, a small mapping effort was needed to create such a system. The mapping between actions in Propara and verbs is straightforward since verbs are automatically mapped onto ontological classes that provide the type of actions based on the parse. Hence, defining the mapping rules for the most general relevant ontology types of verbs is sufficient because all the descendent types will follow the same mapping~(See Table \ref{tab:heuristics}). More details are available in Appendix \ref{appx:propolis}).
Additionally, the effort needed for the pre-processing and designing of the mapping rules is similar to the hyperparameter tuning of neural models. Since mapping is based on common sense rather than trial-and-errors in hyperparameter tuning, finding an optimal solution may even take less effort.

\noindent\textbf{Out-Of-Vocabulary words in parses.}
TRIPS automatically maps words to ontology classes using WordNet~\cite{miller1995wordnet}. This gives us considerable vocabulary coverage and reduces OOV risk. TRIPS can identify the role of the unseen words~(not available in WordNet) based on the sentence syntax and will not produce errors when encountering unseen words. In the same way, PROPOLIS and integrated models will not be affected.

\noindent\textbf{Effectiveness of the new evaluation metric.}
The previously proposed high-level evaluations are strict and do not accurately reflect the quality and quantity of the lower-level model decisions. Thus they do not adequately reveal the models' abilities. For example, when compared at high-level metrics, two models may have the same performance value of $20\%$, while their decision accuracy may be $60\%$ and $10\%$. This issue is reflected during training epochs too when the models' performance remains the same despite the decisions on the train set continuing to improve. Therefore, it seems more appropriate to evaluate the models based on the same objective criteria used for training them~(decision-level). However, the previously used metrics can be secondary evaluations to measure how well the model captures higher-level procedural concepts. 

\section{Conclusion}
We investigated whether semantic parsers could help with reasoning over procedural text. We proposed PROPOLIS, a symbolic model operating on deep semantic parsers to solve the procedural reasoning task. For this task, the symbolic model outperformed many recent neural architectures. We then evaluated the effects of integrating semantic parsers with two well-known SOTA neural backbones. All integrated models outperformed baseline architectures, particularly when the parser provided more detailed information and rich semantic frames. Furthermore, we proposed new evaluation metrics that show the pros and cons of the models and help identify the key challenges in reasoning over procedural text.

\section*{Acknowledgments}
This project is supported by National Science Foundation (NSF) CAREER award 2028626 and partially supported by the Office of Naval Research
(ONR) grant N00014-20-1-2005. Any opinions,
findings, and conclusions or recommendations expressed in this material are those of the authors and do not necessarily reflect the views of the National
Science Foundation nor the Office of Naval Research. We also want to thank Drew Hayward, who helped us with a subset of experiments on PROPOLIS during his research at Michigan State University.

\section*{Limitations}
There are multiple limitations to methods that rely on semantic parsers for solving natural language tasks. First, semantic parsers and especially the ones that do not rely on noisy training data, are most susceptible to errors when the original sentence contains even small grammatical or spelling errors. Next, parsers such as TRIPS rely on general-purpose ontology and a pipeline for generating the output parses. The pipeline first understands the meaning of each word in the sentence. This is subject to errors when words/verbs can have multiple meanings and require the context to disambiguate their semantics. For instance, the TRIPS parser may map the verb `run' to the `management' class in ontology instead of the `physical activity' class. Lastly, executing graph attention networks with many layers requires a powerful system with access to GPU and is more time-consuming than the baselines that do not require reasoning over a graph structure. 
\bibliography{anthology,custom}
\bibliographystyle{acl_natbib}

\appendix
\section{Implementation Details}
\label{appx:imp}
We use the PyTorch geometric~\footnote{\url{https://pytorch-geometric.readthedocs.io/}} library to implement all the graph attention models and Huggingface library~\cite{wolf-etal-2020-transformers} for implementing the language models. For the NCET model and its extensions based on semantic parsers, the best model is selected by a search over the $\lambda \in \{0.3, 0.4\}$, the learning rate in $\{3e-5, 3.5e-5, 5e-5\}$. The number of graph attention layers is set to $2$, and the batch size is set to $8$ process. All models use Bert-base as the selected language model for encoding the context. We further use RAdam~\cite{liu2019variance} to optimize the model parameters of both the language models, the LSTM, and the classifiers.  For the CGLI method, we use the exact hyper-parameters as specified in \cite{ma2022coalescing}. We further use $15$ layers of graph attention network with the input from the fifth layer of the time-aware language models. The gradients from the graph attention network~(GAT) would not back-propagate to the original language model and only affect the parameters in the GAT model.

\section{PROPOLIS}
\label{appx:propolis}
Here, we share more details on the steps in producing symbolic decisions over the actions and the locations of objects in the Propara dataset based on the TRIPS parser. You can also find the ontology of TRIPS parser online\footnote{\url{https://www.cs.rochester.edu/research/trips/lexicon/browse-ont-lex-ajax.html}}.

\subsection{Graph Abstraction}
From the logical forms produced by the TRIPS parser, we need to extract the events and the events' relationships of interest. Because the TRIPS system handles much of the variation expected in sentence constructions, we can use a relatively compact specification for defining the events and relationships of interest while coping with fairly complex and nested formulations.

We capitalized on the TRIPS ontology and parser to develop a compact and easy-to-maintain specification of event extraction rules. Instead of having to write one rule to match each keyword/phrase that could signify an event, many of these words/phrases have already been systematically mapped to a few types in the TRIPS ontology.  For instance, demolish, raze, eradicate, and annihilate are all mapped to the TRIPS ontology type ``ONT::DESTROY''.  In addition, the semantic roles are consistent across different ontology types.  The parser handles various surface structures, and the logical form contains normalized semantic roles.  For example, in the following sentence:
\begin{itemize}
    \item The bulldozer demolished the building
    \item The building was demolished
    \item The demolition of the building
    \item Building demolition
\end{itemize}
, all the parses result in the same basic logical form with the semantic roles ``AFFECTED: the building'' and, where applicable, ``AGENT: the bulldozer''.  Thus, we needed very few extraction rule specifications for each event type, covering a wide range of words and syntactic patterns.

\subsection{Rule-based Local Decisions}
The sets of heuristics used to detect the effect of each semantic frame on the arguments are shown in Table \ref{tab:heuristics}. To handle the location arguments from the parses, we also consider the two cases on `from\_loc' and `to\_loc'. In the specific case of a destroy event, any location attached to the semantic frame is considered the `from\_loc' for the item being destroyed. 

\subsection{Global Reasoning}
To perform the global reasoning over the local predictions, we first do a forward pass through the actions and location predictions and ensure they are globally consistent. To do so, we start from the first predicted action and check the following on every next step prediction:
\begin{itemize}
    \item If the current action is None, then we skip this step!
    \item If the last observed action is ``Create'' or ``Move'',
    \begin{itemize}
        \item If the current action in ``Create'' and the location of this action is the same as the last observed location, then the new ``Create'' action is transformed to ``None''.
        \item IF the current action is ``Create'' and the location of this action is different from the last observed location, then the new action is changed to ``Move''. 
        \item Otherwise, the new action is kept the same, and the last observed action is updated.
    \end{itemize}
    \item If the last observed action is ``Destroy'',
    \begin{itemize}
        \item If the current action is ``Destroy'' and it has a location different from the last observed location, then the action is changed to ``Move''. 
        \item If the current action is ``Destroy'' and it has a location similar to the last observed location, then the action is changed to ``None''. 
        \item Otherwise, the new action is kept the same, and the last observed action is updated.
    \end{itemize}
\end{itemize}
\begin{figure}
    \centering
    \includegraphics[width=\linewidth]{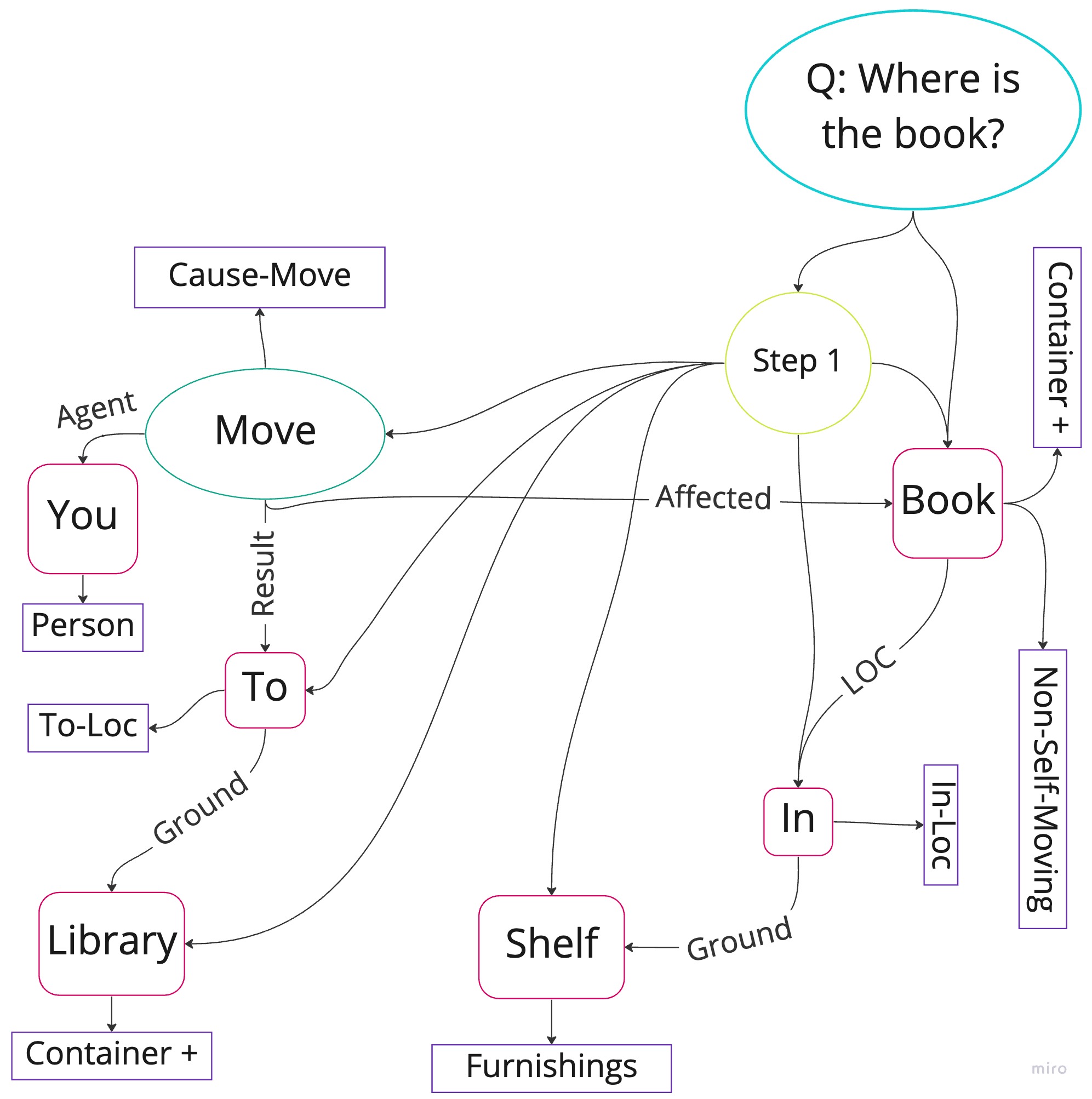}
    \caption{The QA graph for the query of ``where is the book'' and the sentence ``Move the book on the shelf to the library''. }
    \label{fig:qa_graph}
\end{figure}
After fixing the sequence of actions, we first check whether the entity gets created at any of the steps or is just moved or destroyed during the process. If the entity is not created, its initial location is equal to the first `from\_loc' in any subsequent actions.  We then use the following criteria to fix the locations in a forward pass over the local decisions:
\begin{itemize}
    \item If the action is ``Move'' but there is no final location, the final location is the first `from\_loc' from any of the subsequent actions before the next ``Move'' event.
    \item If the object is being ``Moved'', then its final location should be changed. If the action does not indicate a new location or the information is missing, we replace the final location with `?' to indicate an unknown location.
    \item If the action is ``None'', the last location is kept unchanged for the new step. 
\end{itemize}

\section{Graph Attention Network}
\label{appx:gat}
TransformerConv uses the following formula to update the representation of the nodes~($v_i$) in the graph.

\begin{align*}
    \mathbf{v}_i^{l + 1}&=\\&\mathbf{W}_1 \mathbf{v}_i^{l}+\sum_{j \in \mathcal{N}(i)} \alpha_{i, j}\left(\mathbf{W}_2 \mathbf{v}_j^{l}+\mathbf{W}_6 \mathbf{e}_{i j}\right),
\end{align*}
where $\mathcal{N}(i)$ represents the neighbors of node $i$ in the graph, $l$ is the layer, and the coefficient $\alpha_{i, j}$ is computed using the following formula:
\begin{align*}
    &\alpha_{i, j}= \\&{softmax}\left(\frac{\left(\mathbf{W}_3 \mathbf{v}_i^{l}\right)^{\top}\left(\mathbf{W}_4 \mathbf{v}_j^{l}+\mathbf{W}_6 \mathbf{e}_{i j}\right)}{\sqrt{d}}\right)
\end{align*}



\section{Semantic Parsers}
\label{appx:sparser}
Figure \ref{fig:qa_graph} shows an example of the QA graph used in the integration model with the CGLI baseline.

\end{document}